\documentclass[letterpaper, 10 pt, conference]{ieeeconf}  

\IEEEoverridecommandlockouts                              
\usepackage[utf8]{inputenc}
\usepackage{graphicx} 
\usepackage{amsmath} 
\usepackage{amssymb}  
\usepackage{mathtools}
\usepackage{accents}
\usepackage{booktabs}
\usepackage{multirow}
\usepackage{tikz}
\usepackage[mode=buildnew]{standalone}
\usepackage{xcolor, colortbl}
\usepackage{siunitx}
\usepackage{bm}
\usepackage[footnotesize]{caption}
\usepackage[caption=false,font=footnotesize]{subfig}
\usepackage[colorlinks,pagebackref=false,citecolor=blue,bookmarks=false,hypertexnames=true]{hyperref}
\DeclareMathOperator*{\argmax}{argmax}

\definecolor{tu0}{rgb}{0.7451, 0.1176, 0.2353}

\definecolor{tu1}{rgb}{1.0000, 0.8039, 0.0000}
\definecolor{tu11}{rgb}{1.0000, 0.8627, 0.3020}
\definecolor{tu12}{rgb}{1.0000, 0.9020, 0.4980}
\definecolor{tu13}{rgb}{1.0000, 0.9412, 0.6980}
\definecolor{tu14}{rgb}{1.0000, 0.9608, 0.8000}

\definecolor{tu2}{rgb}{0.9804, 0.4314, 0.0000}
\definecolor{tu21}{rgb}{0.9882, 0.6039, 0.3020}
\definecolor{tu22}{rgb}{0.9882, 0.7137, 0.4980}
\definecolor{tu23}{rgb}{0.9922, 0.8275, 0.6980}
\definecolor{tu24}{rgb}{0.9961, 0.8863, 0.8000}

\definecolor{tu3}{rgb}{0.6902, 0.0000, 0.2745}
\definecolor{tu31}{rgb}{0.7529, 0.2000, 0.4196}
\definecolor{tu32}{rgb}{0.8431, 0.4980, 0.6353}
\definecolor{tu33}{rgb}{0.9216, 0.7490, 0.8196}
\definecolor{tu34}{rgb}{0.9529, 0.8510, 0.8902}

\definecolor{tu4}{rgb}{0.4863, 0.8039, 0.9020}
\definecolor{tu41}{rgb}{0.6431, 0.8627, 0.9333}
\definecolor{tu42}{rgb}{0.7412, 0.9020, 0.9490}
\definecolor{tu43}{rgb}{0.8431, 0.9412, 0.9686}
\definecolor{tu44}{rgb}{0.8980, 0.9608, 0.9804}

\definecolor{tu5}{rgb}{0.0000, 0.5020, 0.7059}
\definecolor{tu51}{rgb}{0.3020, 0.6510, 0.7961}
\definecolor{tu52}{rgb}{0.5490, 0.7765, 0.8667}
\definecolor{tu53}{rgb}{0.7490, 0.8745, 0.9255}
\definecolor{tu54}{rgb}{0.8510, 0.9255, 0.9569}

\definecolor{tu6}{rgb}{0.0000, 0.3255, 0.4549}
\definecolor{tu61}{rgb}{0.2510, 0.4941, 0.5922}
\definecolor{tu62}{rgb}{0.5490, 0.6941, 0.7529}
\definecolor{tu63}{rgb}{0.7490, 0.8314, 0.8627}
\definecolor{tu64}{rgb}{0.8510, 0.8980, 0.9176}

\definecolor{tu7}{rgb}{0.7765, 0.9333, 0.0000}
\definecolor{tu71}{rgb}{0.8431, 0.9529, 0.3020}
\definecolor{tu72}{rgb}{0.8863, 0.9647, 0.4980}
\definecolor{tu73}{rgb}{0.9333, 0.9804, 0.6980}
\definecolor{tu74}{rgb}{0.9569, 0.9882, 0.8000}

\definecolor{tu8}{rgb}{0.5373, 0.6431, 0.0000}
\definecolor{tu81}{rgb}{0.6784, 0.7490, 0.3020}
\definecolor{tu82}{rgb}{0.7686, 0.8196, 0.4980}
\definecolor{tu83}{rgb}{0.8588, 0.8941, 0.6980}
\definecolor{tu84}{rgb}{0.9059, 0.9294, 0.8000}

\definecolor{tu9}{rgb}{0.0000, 0.4431, 0.3373}
\definecolor{tu91}{rgb}{0.3020, 0.6118, 0.5373}
\definecolor{tu92}{rgb}{0.5490, 0.7490, 0.7020}
\definecolor{tu93}{rgb}{0.7490, 0.8588, 0.8353}
\definecolor{tu94}{rgb}{0.8549, 0.9176, 0.9059}

\definecolor{tu10}{rgb}{0.8000, 0.0000, 0.6000}
\definecolor{tu101}{rgb}{0.8706, 0.3490, 0.7412}
\definecolor{tu102}{rgb}{0.9216, 0.6000, 0.8392}
\definecolor{tu103}{rgb}{0.9608, 0.8000, 0.9216}
\definecolor{tu104}{rgb}{0.9804, 0.8980, 0.9608}

\definecolor{tu110}{rgb}{0.4627, 0.0000, 0.4627}
\definecolor{tu111}{rgb}{0.5961, 0.2510, 0.5961}
\definecolor{tu112}{rgb}{0.7294, 0.4980, 0.7294}
\definecolor{tu113}{rgb}{0.8392, 0.6980, 0.8392}
\definecolor{tu114}{rgb}{0.9216, 0.8510, 0.9216}

\definecolor{tu120}{rgb}{0.4627, 0.0000, 0.3294}
\definecolor{tu121}{rgb}{0.6118, 0.3020, 0.5333}
\definecolor{tu122}{rgb}{0.7569, 0.5490, 0.6980}
\definecolor{tu123}{rgb}{0.8667, 0.7490, 0.8314}
\definecolor{tu124}{rgb}{0.9216, 0.8510, 0.9020}

\definecolor{tu130}{rgb}{0.0314, 0.0314, 0.0314}
\definecolor{tu131}{rgb}{0.3725, 0.3725, 0.3725}
\definecolor{tu132}{rgb}{0.5882, 0.5882, 0.5882}
\definecolor{tu133}{rgb}{0.7529, 0.7529, 0.7529}
\definecolor{tu134}{rgb}{0.8667, 0.8667, 0.8667}

\definecolor{tu140}{rgb}{0.0000, 0.6875, 0.3125}
\newcommand{\stz}{\rule{0mm}{2.1ex}}

\newcommand{\btb}{     \begin{tabbing}             }
\newcommand{\bte}{     \end{tabbing}               }

\newif\iftitsreview
\newif\ifshownumbers
\titsreviewtrue  
\titsreviewfalse 
\shownumberstrue 
\shownumbersfalse 
\iftitsreview
\ifshownumbers
\newcommand{\revision}[2]{\textcolor{red}{#2 (#1)}}
\else
\newcommand{\revision}[2]{\textcolor{red}{#2}}
\fi
\else
\newcommand{\revision}[2]{#2}
\fi 

\title{\LARGE \bf 
Detecting Adversarial Perturbations in Multi-Task Perception}

\author{Marvin Klingner$^{1}$, Varun Ravi Kumar$^{2}$, Senthil Yogamani$^{2}$, Andreas Bär$^{1}$, and Tim Fingscheidt$^{1}$
\thanks{$^{1}$Marvin Klingner, Andreas Bär, and Tim Fingscheidt are with the Institute for Communications Technology, Technische Universität Braunschweig, Schleinitzstr. 22, 38106 Braunschweig, Germany {\tt\small \{m.klingner, andreas.baer, t.fingscheidt\}@tu-bs.de}}%
\thanks{$^{2}$Varun Ravi Kumar is with Valeo DAR, Kronach, Germany and Senthil Yogamani is with Valeo Vision Systems, Tuam, Ireland.}
}

\begin{document}
 
\maketitle
\thispagestyle{empty}
\pagestyle{empty}

\begin{abstract}
    While deep neural networks (DNNs) achieve impressive performance on environment perception tasks, their sensitivity to adversarial perturbations limits their use in practical applications. In this paper, we (i) propose a novel adversarial perturbation detection scheme based on multi-task perception of complex vision tasks (i.e., depth estimation and semantic segmentation). Specifically, adversarial perturbations are detected by inconsistencies between extracted edges of the input image, the depth output, and the segmentation output. To further improve this technique, we (ii) develop a novel edge consistency loss between all three modalities, thereby improving their initial consistency which in turn supports our detection scheme. We verify our detection scheme's effectiveness by employing various known attacks and image noises. In addition, we (iii) develop a multi-task adversarial attack, aiming at fooling both tasks as well as our detection scheme. Experimental evaluation on the Cityscapes and KITTI datasets shows that under an assumption of a 5\% false positive rate up to 100\% of images are correctly detected as adversarially perturbed, depending on the strength of the perturbation. Code is available at \href{https://github.com/ifnspaml/AdvAttackDet}{\url{https://github.com/ifnspaml/AdvAttackDet}}. A short video at {\url{https://youtu.be/KKa6gOyWmH4}} provides qualitative results.
\end{abstract}
 
\section{Introduction}
\label{sec:introduction}

\begin{figure}[t]
	\centering
	\vspace{0.1cm}
	\includegraphics[width=1.0\linewidth]{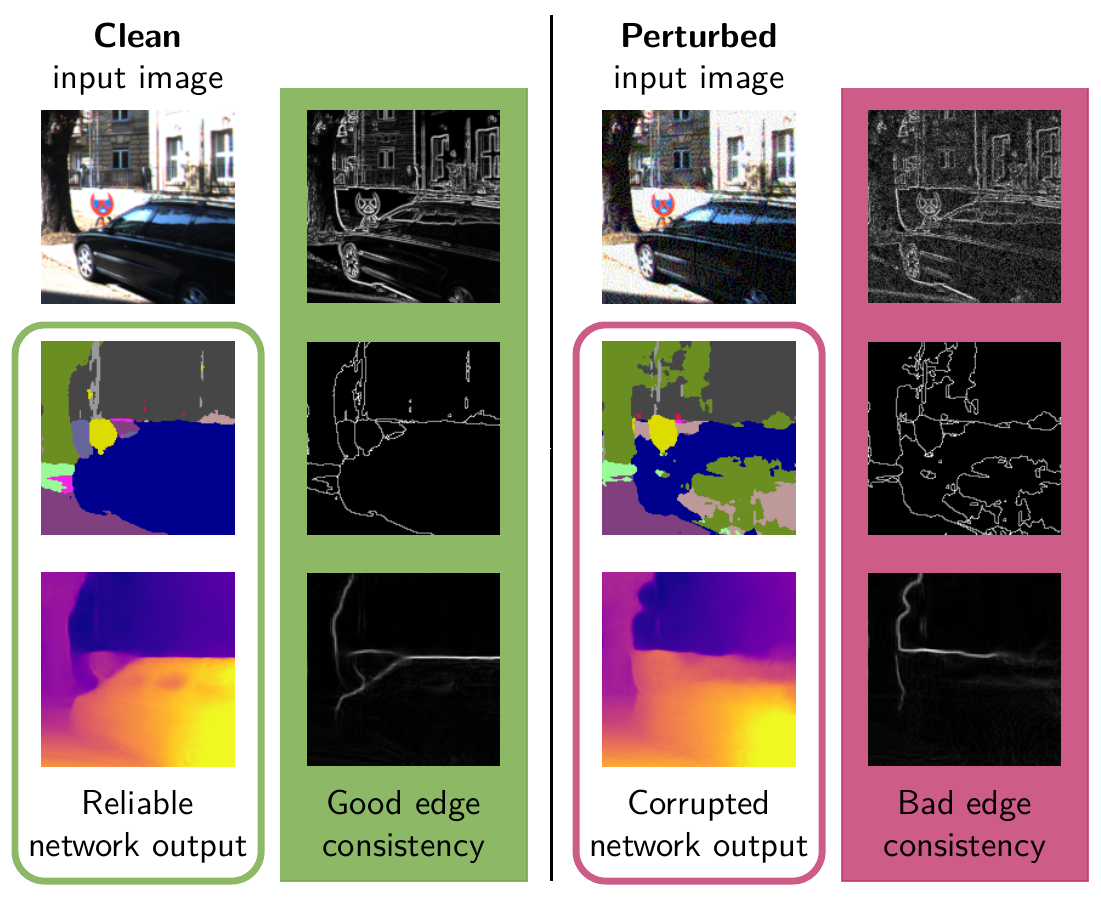}
	\caption{\revision{E.1, E.6}{\textbf{Motivation} for} our detection framework. Left-right pairs of input images and edges (top row), segmentation outputs and edges (middle row), and depth outputs and edges (bottom row) are depicted. A clean input image (left half) has consistent edges with the depth and segmentation network outputs. If the input image is perturbed (right half), significantly reduced input-output and output-output edge consistencies can be observed.}
	\label{fig:high_level_concept}
	\vspace{-0.5cm}
\end{figure}

\textbf{Motivations}:
\revision{Motivation for real-time detection}{Recent research has shown that on input images to deep neural networks (DNNs) visually imperceptible perturbations exist \cite{Goodfellow2015, Kurakin2017} and lead to network performance drops (cf.~Fig.~\ref{fig:high_level_concept}). As various of these perturbations generalize across different images and network architectures \cite{Moosavi-Dezfooli2017} or are implemented as patches which can simply be pasted on cameras \cite{Nesti2022}, they are an imminent danger in safety-critical applications including autonomous driving. A possible mitigation of this problem is to detect failures caused by these perturbations (or other out-of-distribution samples) in real-time, i.e., adversarial perturbation detection. Subsequently, high-level planning systems can be warned about the unreliability of the network output (cf.~Fig.~\ref{fig:high_level_concept}) such that measures can be taken accordingly.}
\par 
\textbf{Detecting Adversarial Perturbations}:
\revision{Differences to image classification}{As a result of adversarial perturbations' discovery, many works emerged, proposing methods for improved robustness towards these perturbations \cite{Hendrycks2019a, Madry2018}, as well as adversarial perturbation detection methods \cite{Hendrycks2017, Liu2019g, Tian2021, Sperl2020}. Current works in this field usually focus on image classification, as this task is simple and well understood. For example, they use additional detection networks \cite{Liu2019g}, analyze the network output~\cite{Hendrycks2017, Tian2021}, or make use of certain activation patterns in the hidden layers~\cite{Sperl2020}. These detection approaches are often specific to the output structure and/or network topology of image classification and are thereby not easily transferable to more complex vision tasks. Accordingly, for our addressed tasks of semantic segmentation and depth estimation only few single-task approaches exist \cite{Baer2020, Mao2020}. In this work, we are the first to use the more complex output structure of a multi-task network to implement an adversarial perturbation detection, i.e., we make use of three pair-wise edge consistencies between the input image and the two network outputs (cf.~Fig.~\ref{fig:high_level_concept}). If two out of three edge consistencies drop below thresholds, our method indicates an adversarial perturbation. This also implements a redundant detection mechanism such that if one pair-wise edge inconsistency is not detected, the other two edge inconsistencies would still be enough to detect the adversarial perturbation. Also, our detector can be appended to a given multi-task DNN in a straightforward manner.}
\par
\textbf{Adversarial Perturbations for Multi-Task Networks}:
Environment perception systems are often implemented as multi-task networks, trained by multi-task learning (MTL) \cite{Kendall2018}. \revision{Differences to baselines and to [28]}{Here, MTL and edge consistency is used to improve performance and efficiency \cite{Klingner2020a, Liu2019a, Yang2018c, Chen2019a, Zhu2020}. In contrast, we propose an edge consistency loss (ECL) to improve adversarial perturbation detection. Some prior works also enforce edge consistency between the depth and segmentation outputs  \cite{Chen2019a, Zhu2020}. However, to the best of our knowledge, we are the first to employ edge consistency for all three possible pair-wise consistencies between RGB image input, depth output, and segmentation output, both as loss during training and as metric to detect perturbations. Furthermore}, Mao~et~al.~\cite{Mao2020} have shown that MTL strengthens adversarial robustness due to the increased difficulty of successfully attacking several tasks\revision{theoretical analysis}{, which is also supported by a theoretical analysis.} Similarly, subsequent works aimed at exploring different task combinations \cite{Klingner2020, Wang2020c, RaviKumar2021}, comparing the effectiveness of adding different auxiliary tasks \cite{Ghamizi2021}, or studying the relatedness of tasks by perturbation generalizability~\cite{Gurulingan2021, Haleta2021}. While the positive effects of MTL on adversarial \textit{robustness} are quite well explored, we are the first to propose an adversarial perturbation \textit{detection} based on MTL of semantic segmentation and depth estimation.
\par
\textbf{Contributions}: 
In this work, we propose an adversarial perturbation detection mechanism based on edge extraction from the input image, the semantic segmentation output, as well as from the depth output. By comparing their edge consistency as shown in Fig.\ \!\ref{fig:high_level_concept}, we can distinguish clean images with good edge consistency from perturbed images with rather bad edge consistency. \revision{Focus on novelty of the detection for the editor}{To the best of our knowledge, we are the first to use edge consistency both as loss and metric for adversarial perturbation detection.} Specifically, our contributions are as follows: First, we address adversarial perturbation detection for the tasks of depth estimation and semantic segmentation. Second, we propose a detection method based on edge consistencies of the input image and the depth and segmentation outputs of a multi-task network. Third, we propose a consistency loss function to further support our detection method. Fourth, we verify our detection method against an attacker with white-box knowledge of both the multi-task network \textit{and} our detection method. Finally, we show the effectiveness of our detection method against several adversarial attacks, in particular the Orthogonal-PGD (O-PGD) attack~\cite{Bryniarski2022} that recently fooled most state-of-the-art adversarial perturbation detection methods on the image classification task.

\section{Training and Detection Method Description}
\label{sec:method_description}

In this section we outline our edge-consistent multi-task learning (MTL) framework (cf.~Fig.~\ref{fig:training_concept}) as well as our adversarial perturbation detection method (cf.~Fig.~\ref{fig:detection_concept}).

\subsection{Edge-Consistent Multi-Task Learning Framework}
\label{sec:multi_task_learning}

\begin{figure}[t]
	\centering
	\vspace{0.2cm}
	\includegraphics[width=1.0\linewidth]{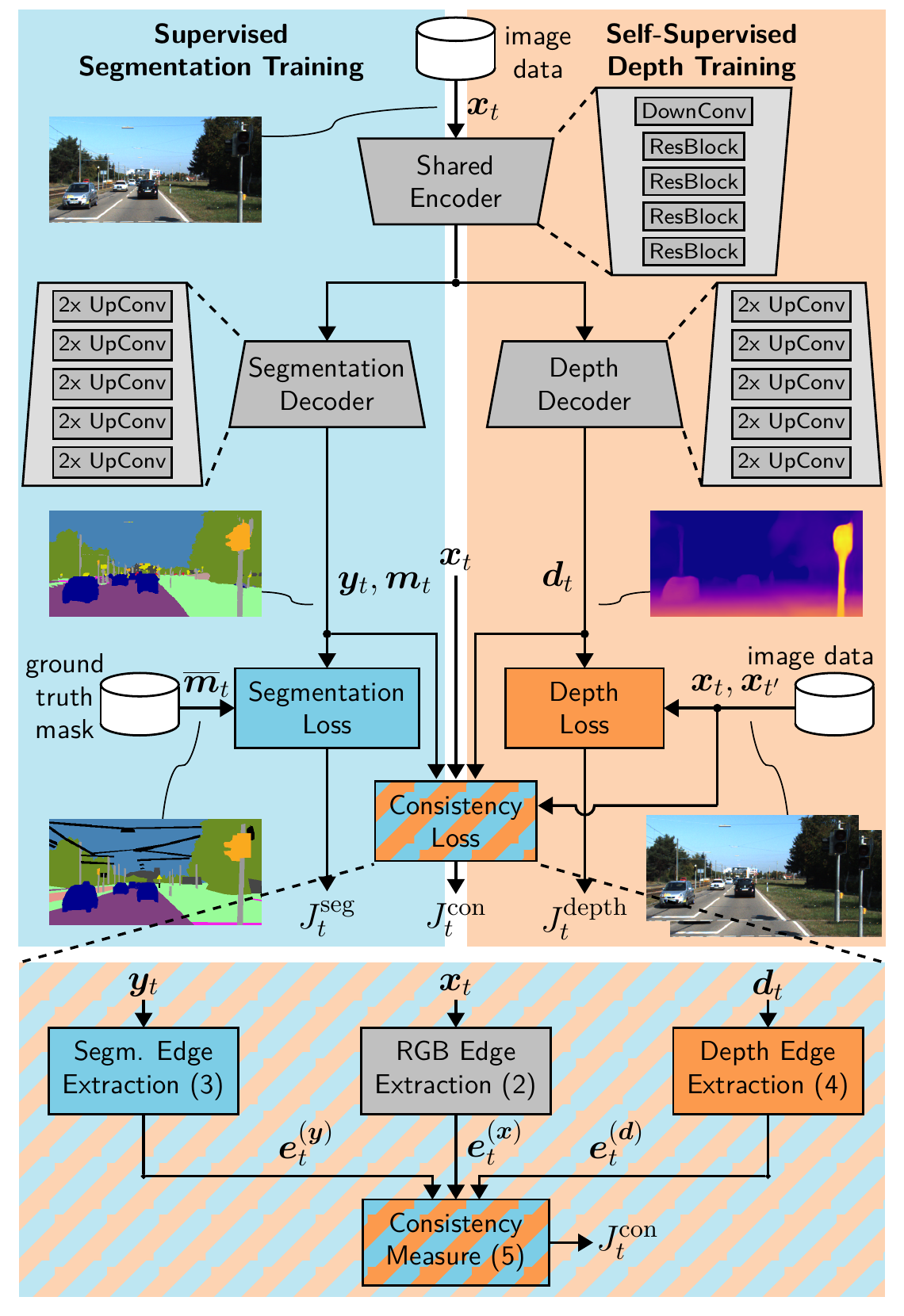}
	\vspace{-0.5cm}
	\caption{\textbf{Edge-consistent multi-task training}: The single tasks of the multi-task network are trained by their respective losses. Additionally, we apply our edge-consistency loss between the RGB image and the network outputs.}
	\label{fig:training_concept}
	\vspace{-0.5cm}
\end{figure}

\textbf{Multi-Task Learning Setup}: 
For joint estimation of depth and segmentation we employ a single encoder and two task-specific decoders as shown in Fig.~\ref{fig:training_concept}. The input to our network is a single RGB image $\bm{x}_t = (x_{t,i,c})\in\mathbb{I}^{H\times W\times C}$ at time instance $t$ with height $H$, width $W$, $C=3$ color channels, and pixel position $i\in \mathcal{I} = \left\lbrace 1, ..., H\cdot W\right\rbrace$ normalized to the range $\mathbb{I} = \left[0,1\right]$. The segmentation decoder outputs class probabilities $\bm{y}_t = (y_{t,i, s}) \in\mathbb{I}^{H\times W\times |\mathcal{S}|}$ for $|\mathcal{S}|$ semantic classes. The class probabilities are normalized pixel-wise over the set of classes $\mathcal{S} = \left\lbrace 1, ..., |\mathcal{S}|\right\rbrace$ such that $\sum_{s\in\mathcal{S}} y_{t,i, s} = 1$. Accordingly, the final classes $\bm{m}_t = (m_{t,i})\in \mathcal{S}^{H\times W}$ can be obtained pixel-wise  by $m_{t,i} = \argmax_{s \in \mathcal{S}} y_{t, i, s}$. The depth decoder outputs a depth map $\bm{d}_t = (d_{t,i}) \in \mathbb{D}^{H\times W}$ with pixel-wise depth values $d_{t,i}$ in the pre-defined depth range $\mathbb{D} = \left[d^{\mathrm{min}},d^{\mathrm{max}}\right]$ with lower bound $d^{\mathrm{min}}$ and upper bound $d^{\mathrm{max}}$. 
\par
The network is trained using task-specific losses $J_t^{\mathrm{depth}}$ and $J_t^{\mathrm{seg}}$ for depth and segmentation, respectively. \revision{Differences to [28] and shorten}{We use the exact same network architecture and single-task loss definitions as~\cite{Klingner2020a} but do not make use of the dynamic-object masking as it was not beneficial for our adversarial perturbation detection approach. The segmentation loss $J_t^{\mathrm{seg}}$ is a class-balanced cross-entropy loss between the class probabilities $\bm{y}_t$ and the one-hot encoded ground truth labels $\overline{\bm{y}}_t = \left(\overline{y}_{t,i,s}\right)\in \left\lbrace 0,1\right\rbrace^{H\times\ W\times |\mathcal{S}|}$, while the depth loss $J_t^{\mathrm{depth}}$ is a reprojection error between pairs of consecutive images ($\bm{x}_t$, $\bm{x}_{t'}$) with $t'\in\mathcal{T}' = \left\lbrace {t\!-\!1}, {t\!+\!1} \right\rbrace$. More loss details can be found in \cite{Klingner2020a}. Finally, (and in contrast to \cite{Klingner2020a}) we employ} our novel edge consistency loss $J^{\mathrm{con}}$, yielding the total loss
\begin{equation}
    J_t^{\mathrm{tot}} = J_t^{\mathrm{seg}} + J_t^{\mathrm{depth}} + \mu J_t^{\mathrm{con}}.
    \label{eq:total_loss}
\end{equation}
While $\mu$ regularizes the influence of the edge consistency loss, the influence of the single tasks is weighted by a gradient weighting factor $\lambda = 0.1$ in the encoder \cite{Ganin2015, Klingner2020a}.
\par
\textbf{Novel Edge-Consistency Training}: 
As we detect perturbations by low edge consistency of network input and outputs, we aim at increasing the edge consistency between them on unperturbed images by a novel edge consistency loss. Specifically, we apply edge consistency losses between the three pairs $(k, \kappa)\in \mathcal{K} = \left\lbrace (\bm{y}, \bm{x}), (\bm{x}, \bm{d}), (\bm{y}, \bm{d}) \right\rbrace$. As a first step, we extract their edges. For the RGB image, we extract edges for pixel position $i$ as
\begin{equation}
\bm{e}_{t, i}^{(\bm{x})} = (e_{t, i, n}^{(\bm{x})}) = \left(\frac{1}{C}\left\lVert\partial_\mathrm{h} \bm{x}_{t,i}\right\rVert_1, \frac{1}{C}\left\lVert\partial_\mathrm{w} \bm{x}_{t,i}\right\rVert_1 \right)\!\in\!\mathbb{R}_+^2,
\label{eq:rgb_edges}
\end{equation}
by calculating the gradients (i.e., the value difference) $\partial_n$ between neighboring pixels in height ($n=\mathrm{h}$) and width dimension ($n=\mathrm{w}$). Similarly, the segmentation edges 
\begin{equation}
\bm{e}_{t, i}^{(\bm{y})} = (e_{t, i, n}^{(\bm{y})}) = \left(\frac{1}{|\mathcal{S}|}\left\lVert\partial_\mathrm{h} \bm{y}_{t,i}\right\rVert_1, \frac{1}{|\mathcal{S}|}\left\lVert\partial_\mathrm{w} \bm{y}_{t,i}\right\rVert_1 \right)\!\in\!\mathbb{R}_+^2,
\label{eq:seg_edges}
\end{equation}
are obtained from the class probabilities $\bm{y}_{t}$ as the actual segmentation map $\bm{m}_{t}$ is obtained by a non-differentiable argmax operation and can therefore not be used during training. For the depth output, we extract edges 
\begin{equation}
\bm{e}_{t, i}^{(\bm{d})} = (e_{t, i, n}^{(\bm{d})}) = \left(|\partial_\mathrm{h} \tilde{\eta}_{t,i}|, |\partial_\mathrm{w} \tilde{\eta}_{t,i}| \right)\!\in\!\mathbb{R}_+^2
\label{eq:depth_edges}
\end{equation}
from the mean-normalized inverse depth $\tilde{\bm{\eta}}_t\in \mathbb{R}^{H\times W}$, whose elements are computed by $\tilde{\eta}_{t,i} = \frac{\eta_{t,i}}{\frac{1}{HW} \sum_{j\in\mathcal{I}} \eta_{t,j}}$ with $\eta_{t,i} = \frac{1}{d_{t,i}}$, as previous works usually use this formulation for a first-order depth gradient in a loss function~\cite{Godard2019, Casser2019}. 
\par
Our proposed edge-consistency loss encourages edges of modality $\kappa$ at pixels $i$ to be structurally similar to edges of modality $k$ which we achieve using the SSIM difference. \revision{motivation of SSIM}{We choose this difference since edges can be depicted as images and the SSIM difference is commonly used to compare the similarity of images. Accordingly, our loss is defined as}
\begin{equation}
    J_t^{\mathrm{con}} = 1\!-\!\frac{1}{6|\mathcal{I}|}\sum_{i \in\mathcal{I}}\sum_{(k,\kappa)\in\mathcal{K}} \sum_{n\in\left\lbrace \mathrm{h},\mathrm{w}\right\rbrace}
    \mathrm{SSIM}_i\!\left( \bm{e}_{t, n}^{(\kappa)}, \bm{e}_{t, n}^{(k)} \right)\!. 
    \label{eq:edge_consistency_loss}
\end{equation}
We average over the SSIM difference of $3\times3$ patches around pixel index $i\in\mathcal{I}$, modality pairs $(k,\kappa)\in\mathcal{K}$, and gradient directions $n\in\left\lbrace \mathrm{h},\mathrm{w} \right\rbrace$ (cf.~(\ref{eq:rgb_edges}), (\ref{eq:seg_edges}), and (\ref{eq:depth_edges})). In other words, (\ref{eq:edge_consistency_loss}) can be interpreted in a way that segmentation and depth edges are encouraged at pixels with high RGB value gradients and depth edges are further encouraged in regions with high segmentation class probability gradients. Our loss (\ref{eq:edge_consistency_loss}) may not be optimal for performance due to shadows or textures (in which neither depth nor segmentation edges are desired) but it rather aims at increasing the sensitivity of the adversarial perturbation detection.

\subsection{Adversarial Perturbation Detection by Edge Consistency}
\label{sec:detection_method}

\begin{figure*}[t]
	\centering
	\vspace{0.15cm}
	\includegraphics[width=1.0\linewidth]{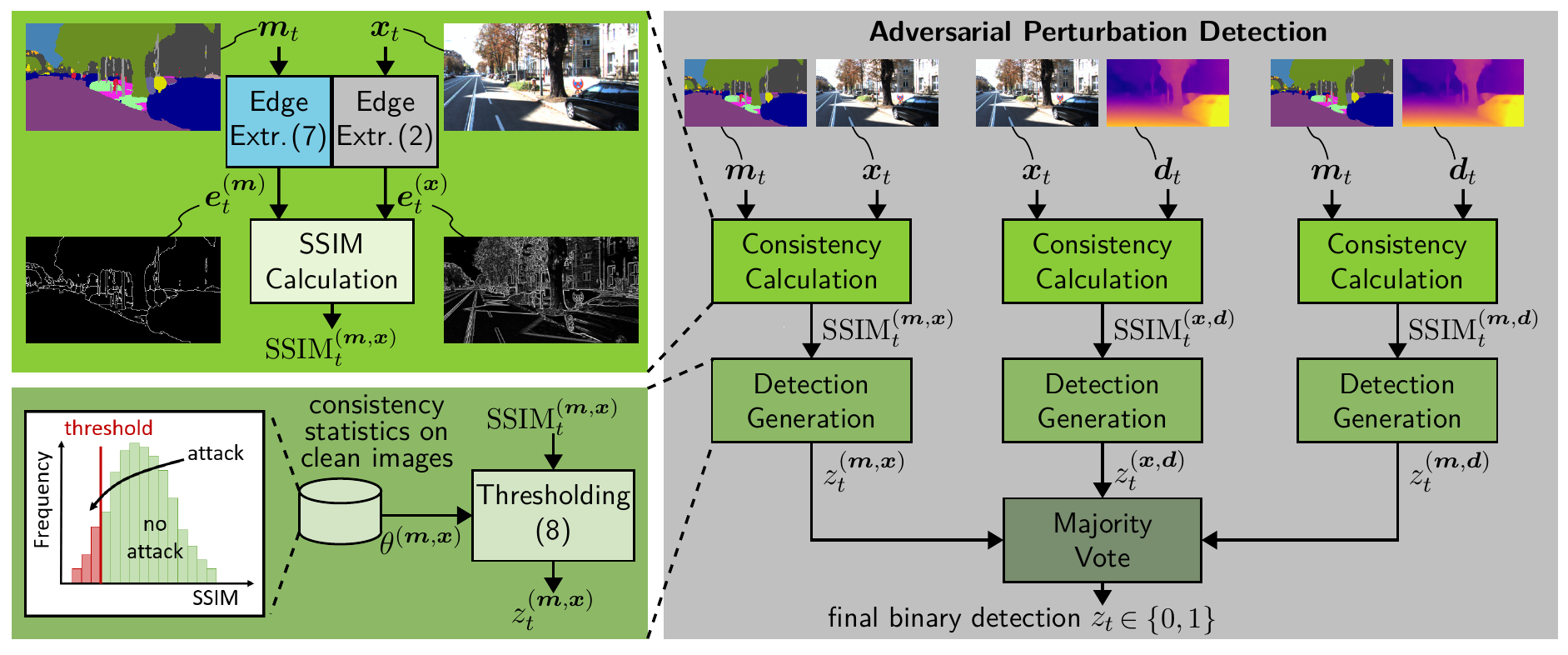}
	\vspace{-0.4cm}
	\caption{\textbf{Overview of our adversarial perturbation detection method} for semantic segmentation and depth estimation. We first calculate the three edge consistencies $\mathrm{SSIM}^{(k,\kappa)}_t$ pairwise between the (perturbed) input image, the depth output and the segmentation output. Afterwards, the binary detection output is generated by comparing any of the three edge consistencies to a threshold $\theta^{(k,\kappa)}$. The final detection output is created by majority vote.}
	\vspace{-0.6cm}
	\label{fig:detection_concept}
\end{figure*} 

The aim of our adversarial perturbation detection method depicted in Fig.~\ref{fig:detection_concept} is to detect input perturbations $\bm{r}_{\epsilon}\in \left[-1,1\right]^{H\times W\times C}$ of strength $\epsilon= \sqrt{\frac{1}{HWC}\operatorname{E}\big(\left\lVert\bm{r}_\epsilon \right\rVert^2_2\big)}$ (defined by the expectation value of the sum of the squared noise pixels) on an input image $\bm{x}_{t}$, see also \cite{Klingner2020}. The perturbation is added, yielding the perturbed image 
\begin{equation}
	\bm{x}_{t,\epsilon} = \bm{x}_t + \bm{r}_{t, \epsilon}.
	\label{eq:add_input_perturbation}
\end{equation}
For simplicity and consistency to Sec.~\ref{sec:multi_task_learning}, we will omit the index $\epsilon$ in the following, i.e., $\bm{x}_{t,\epsilon} \widehat{=}\, \bm{x}_{t}$.
\par
\textbf{Edge Consistency Calculation}:
As shown in Fig.~\ref{fig:detection_concept}, the inputs to our detection method are the (perturbed) input image $\bm{x}_{t}$, the depth output $\bm{d}_{t}$, and the segmentation output $\bm{m}_{t}$. As a first step, we extract the edges of all three modalities. \revision{motivation of edge definition}{Note that we aim at extracting edges, suitable for adversarial perturbation detection, and not at improving the fidelity of the edges.} The RGB image edges and the depth edges are extracted according to (\ref{eq:rgb_edges}) and (\ref{eq:depth_edges}), respectively. Regarding the segmentation map, an edge is simply defined as two neighboring pixels containing different classes. Accordingly, the segmentation edges $\bm{e}_{t}^{(\bm{m})}$ are computed by 
\begin{equation}
\bm{e}_{t, i}^{(\bm{m})} = \left( \left[ |\partial_\mathrm{h} m_{t,i}| > 0 \right] , \left[ |\partial_\mathrm{w} m_{t,i}| > 0 \right] \right)\in\left\lbrace0,1\right\rbrace^2
\label{eq:seg_edges_binary}
\end{equation}
where $\left[\cdot\right]$ is the Iverson bracket, which yields $1$, if the condition in the bracket is true, and $0$ otherwise. We then calculate three pairwise consistencies between $\bm{e}_{t}^{(k)}, \bm{e}_{t}^{(\kappa)}$ with $(k,\kappa)\in\mathcal{K}' = \left\lbrace (\bm{m}, \bm{x}), (\bm{x}, \bm{d}), (\bm{m}, \bm{d}) \right\rbrace$ using the SSIM metric defined in \cite{wang2004image}, yielding the edge consistencies $\mathrm{SSIM}^{(k,\kappa)}_t = \mathrm{SSIM}\big(\bm{e}_{t}^{(k)}, \bm{e}_{t}^{(\kappa)}\big)$ as shown in the top left part of Fig.~\ref{fig:detection_concept}. Note that in contrast to the edge-consistency \textit{loss}, the structural similarities $\mathrm{SSIM}^{(k,\kappa)}_t$ of the detector are computed according to the original formulation \cite{wang2004image}, and in non-differentiable fashion because the segmentation map $\bm{m}_t$ is computed using the argmax operation, making a direct calculation of adversarial perturbations impossible.
\par
\textbf{Detection Generation}:
To generate a detection score $z_t^{(k,\kappa)}$, denoting whether an input image is adversarially perturbed, we further rely on a threshold $\theta^{(k,\kappa)}$ (cf.~bottom left part in Fig.~\ref{fig:detection_concept}). This threshold is obtained from the distribution of consistencies on the validation set where images are known to be unperturbed. We choose the threshold such that a fraction $\gamma\in\left[0,1\right]$ of consistencies is below the threshold. If $\gamma=0$, the detector never wrongfully classifies clean images as perturbed. If the edge consistency is below the threshold $\theta^{(k,\kappa)}$, it indicates the presence of an adversarial perturbation. Mathematically, the detection result $z_t^{(k,\kappa)}$ is obtained as
\begin{equation}
z_t^{(k,\kappa)} = \big[\mathrm{SSIM}\big(\bm{e}_{t}^{(k)}, \bm{e}_{t}^{(\kappa)}\big) < \theta^{(k,\kappa)} \big],
\label{eq:detection_single}
\end{equation}
where $\left[\cdot\right]$ again represents the Iverson bracket. The final detection $z_t$ is generated by using a majority vote of the three single detections $z_t^{(\bm{m},\bm{x})}$, $z_t^{(\bm{x},\bm{d})}$, and $z_t^{(\bm{m},\bm{d})}$ as shown in the bottom right part of Fig.~\ref{fig:detection_concept}.

\subsection{Multi-Task Adversarial Attack Design}

An untargeted adversarial attack aims at optimizing the perturbation $\bm{r}_{t,\epsilon}$ in (\ref{eq:add_input_perturbation}) to minimize DNN performance. Accordingly, the negative impact of an adversarial perturbation is usually stronger compared to random perturbations. For an in-depth overview we report on Gaussian noise and salt and pepper (S\&P) noise, representing random perturbations, and on FGSM~\cite{Goodfellow2015}, BIM~\cite{Kurakin2017}, PGD~\cite{Madry2018}, and O-PGD~\cite{Bryniarski2022}, representing adversarial perturbations. We apply all attacks with the standard settings reported in the respective publications.
\par
\textbf{Adversarial Perturbation Calculation}:
In the following, we specify a new attack aiming at fooling both modalities of our multi-task perception as well as our detection method. We exemplarily describe its calculation for the FGSM adversarial attack \cite{Goodfellow2015}, while its extension to more complex attacks is achieved straightforward by the iterative computation protocol used in such attacks.
\par
For the FGSM attack, the perturbation is calculated by
\begin{equation}
    \bm{r}_{t, \epsilon} =  \epsilon \cdot \mathrm{sign}\left(\bm{\nabla}_{\bm{x}} J_t^{\mathrm{adv}}\right),
    \label{eq:perturbation_calculation_fgsm}
\end{equation}
where we calculate the derivative $\bm{\nabla}_{\bm{x}}$ of our adversarial perturbation loss function $J_t^{\mathrm{adv}}$ with respect to the input image $\bm{x}_t$ and apply the element-wise signum function $\mathrm{sign}(\cdot)$. Accordingly, the adversarial perturbation $\bm{r}_{t,\epsilon}$ is optimized to maximize $J_t^{\mathrm{adv}}$ by (\ref{eq:perturbation_calculation_fgsm}). While previous works usually optimize attacks for a single task, we aim at fooling both tasks of the multi-task network by setting
\begin{equation}
    J_t^{\mathrm{adv}} = J_t^{\mathrm{depth}} + J_t^{\mathrm{seg}},
    \label{eq:multi_task_attack}
\end{equation}
with $J_t^{\mathrm{depth}}$ and $J_t^{\mathrm{seg}}$ computed as in~\cite{Klingner2020a}, respectively. Maximizing (\ref{eq:multi_task_attack}) by (\ref{eq:perturbation_calculation_fgsm}) naturally induces a lower performance. 
\par
\textbf{Incorporating the Detection Method}:
If the attacker calculating the adversarial perturbation $\bm{r}_{t,\epsilon}$ has knowledge about the detection, he could additionally aim at fooling the detection method. As our detection relies on inconsistency between input and outputs, the attacker could additionally aim at achieving a high edge consistency, characterized by a low edge consistency loss $J_t^{\mathrm{con}}$ (\ref{eq:edge_consistency_loss}). Accordingly, the adversarial perturbation loss can be written as
\begin{equation}
    J_t^{\mathrm{adv}} = J_t^{\mathrm{depth}} + J_t^{\mathrm{seg}} - \tilde{\mu} J_t^{\mathrm{con}},
    \label{eq:multi_task_attack_detection}
\end{equation}
where $\tilde{\mu}$ is a loss weighting factor.
This objective is difficult to solve for an attacker, as optimizing the adversarial perturbation $\bm{r}_{t,\epsilon}$ towards a better consistency during one training step may make the image $\bm{x}_{t,\epsilon}$ (cf.~(\ref{eq:add_input_perturbation})) inconsistent to the outputs of the subsequent training step.

\section{Implementation Details}
\label{sec:implementation_details}

\begin{table*}[t]
  \centering
  \vspace{0.15cm}
  \caption{\textbf{Prediction performance ablation and comparison to baselines} on the KITTI-2015 validation set. We show depth and segmentation performance as well as the consistency between the RGB input image and the depth and segmentation outputs. We compare the segmentation and the depth single-task models, a multi-task (MT) model, a model trained with an additional smoothness loss~\cite{Godard2017}, models trained with our edge consistency loss (ECL) (\ref{eq:edge_consistency_loss}) but only one pair being evaluated: $(k,\kappa) = (\bm{y}, \bm{x})$ or $(k,\kappa) = (\bm{x}, \bm{d})$ or $(k,\kappa) = (\bm{y}, \bm{d})$, and our proposed model trained with the full ECL of (\ref{eq:edge_consistency_loss}) (\textbf{Ours}). \revision{monocular vs stereo}{Best monocular results in boldface (as the stereo result should be better by nature)}, second-best underlined. \revision{arrow explanation}{Arrows indicate if high or low values are better.}}
  \resizebox{\textwidth}{!}{
  \setlength{\tabcolsep}{2.5pt}
  \begin{tabular}{l|c|ccccccc|ccc}
    & Segmentation & \multicolumn{7}{c|}{Depth} & \multicolumn{3}{c}{Consistency}\\
    \textbf{Method} & $\mathrm{mIoU}\left[\%\right]$\,$\uparrow$ & Abs Rel\,$\downarrow$ & Sq Rel\,$\downarrow$ & RMSE\,$\downarrow$ & RMSE log\,$\downarrow$ & $\delta < 1.25$\,$\uparrow$ & $\delta < 1.25^2$\,$\uparrow$ & $\delta < 1.25^3$\,$\uparrow$ & $\overline{\mathrm{SSIM}}^{(\bm{m},\bm{x})}$\,$\uparrow$ & $\overline{\mathrm{SSIM}}^{(\bm{x},\bm{d})}$\,$\uparrow$ & $\overline{\mathrm{SSIM}}^{(\bm{m},\bm{d})}$\,$\uparrow$\\
    \hline  
    \revision{red}{Zhu~et~al.~\cite{Zhu2020} (stereo)} & - & \revision{red}{0.097} & \revision{red}{0.675} & \revision{red}{4.350} & \revision{red}{0.180} & \revision{red}{0.890} & \revision{red}{0.964} & \revision{red}{0.983} & - & - & - \\
    \hline 
    EPC~\cite{Yang2018c} & - & 0.131 & 1.254 & 6.117 & 0.220 & 0.826 & 0.931 & 0.973 & - & - & - \\
    UnRigidFlow~\cite{Liu2019a} & - & 0.108 & 1.020 & 5.528 & 0.195 & 0.863 & 0.948 & 0.980 & - & - & - \\
    SGDepth~\cite{Klingner2020a} & 50.1 & \textbf{0.097} & \textbf{0.983} & 6.173 & \textbf{0.160} & \textbf{0.898} & 0.972 & \textbf{0.990} & \revision{red}{\underline{0.20}} & \revision{red}{0.34} & \revision{red}{\underline{0.44}} \\
    \hline
    only segmentation & 45.2 & - & - & - & - & - & - & - & \revision{red}{\underline{0.20}} & - & - \\
    only depth & - & 0.107 & 1.082 & 6.470 & 0.173 & 0.878 & 0.967 & \underline{0.988} & - & \revision{red}{0.34} & - \\
    Ours w/o ECL = MT & 50.1 & \underline{0.099} & \underline{0.992} & 6.219 & \underline{0.162} & \underline{0.894} & 0.971 & \textbf{0.990} & \underline{0.20} & 0.34 & 0.41 \\
    MT + smooth~\cite{Godard2017} & 47.4 & 0.101 & 0.997 & \textbf{6.125} & 0.163 & 0.893 & 0.972 & \textbf{0.990} & \underline{0.20} & 0.34 & 0.43 \\
    MT + ECL$(\bm{y}, \bm{x})$ & 48.8 & 0.102 & 1.037 & 6.149 & 0.163 & \underline{0.894} & \textbf{0.973} & \textbf{0.990} & \underline{0.20} & 0.34 & \underline{0.44} \\
    MT + ECL$(\bm{x}, \bm{d})$ & \underline{51.1} & 0.106 & 1.109 & 6.285 & 0.168 & 0.885 & 0.971 & \textbf{0.990} & \underline{0.20} & \textbf{0.52} & 0.24 \\
    MT + ECL$(\bm{y}, \bm{d})$ & 48.5 & 0.103 & 1.034 & 6.178 & 0.164 & 0.891 & 0.971 & \textbf{0.990} & \underline{0.20} & 0.33 & \textbf{0.50} \\
    \hline
    \textbf{Ours}\stz & \textbf{51.6} & 0.101 & 1.030 & \underline{6.137} & \underline{0.162} & \underline{0.894} & \underline{0.972} & \textbf{0.990} & \textbf{0.21} & \underline{0.36} & \underline{0.44} \\
  \end{tabular}}
  \vspace{-0.3cm}
  \label{tab:prediction_performance}
\end{table*}

Here, we describe our used datasets, evaluation metrics, network architecture, and multi-task training details to facilitate reproducibility. We employ the \texttt{PyTorch} library~\cite{Paszke2019} executed on an \texttt{NVIDIA Tesla P100} graphics card.
\par
\textbf{Datasets}:
We use the well-established benchmark datasets KITTI~\cite{Geiger2013} and Cityscapes~\cite{Cordts2016}. The KITTI dataset is generally used to benchmark depth estimation models. We utilize the 29,000 images defined by~\cite{Godard2017} for training as well as 200 images from the KITTI-2015 dataset~\cite{Menze2015} for validation. On the other hand, the Cityscapes dataset is generally used to benchmark semantic segmentation models for urban scene understanding. It contains well-defined splits of 2,975 and 500 images for training and validation, respectively. \revision{dataset usage}{The multi-task training was carried out on the Cityscapes and KITTI training datasets. After training, we took design decisions regarding our detection method on the KITTI validation set (cf.~Table~\ref{tab:ablation_detection}), subsequently testing it without any changes on the Cityscapes validation set to verify our method's generalizability to unknown data.}
\par
\textbf{Evaluation Metrics}:
We evaluate the semantic segmentation task using the mean intersection over union (mIoU) metric as defined in~\cite{Everingham2015}. For the depth estimation task, we use the commonly reported four error metrics as well as three accuracy metrics as defined in~\cite{Eigen2014}. To evaluate our detection method, we follow common practices\footnote{We followed \cite{Liu2019g} in reporting the detection rate as it is easier to interpret compared to the F1 or AUC score and allows for a good comparison of different detectors through a common fixed false positive rate. Note that the F1 score can be obtained in straightforward fashion from the fixed false positive rate and the reported detection rate.} in using the detection success rate $\mathrm{TPR}_{5\%}$ (true positive rate) while fixing the false positive rate (FPR) on clean images to 5\% through the thresholds $\theta^{(k,\kappa)}$ (cf.~Sec.~\ref{sec:detection_method}). We additionally report the prediction performance of the multi-task DNN under perturbations, which is also of interest, since a high detection rate is only important at a low prediction performance.
\par
\textbf{Network Architecture}:
\revision{network architecture}{As can be seen in Fig.~\ref{fig:training_concept}, we employ an encoder-decoder architecture. The input image is first passed through a \texttt{ResNet-18} encoder adopted from~\cite{He2016}, which consists of a convolutional layer and four residual blocks (each block contains four convolutional layers with residual connections). Afterwards, the extracted features are passed through task-specific decoders with identical architecture whose design is adopted from \cite{Klingner2020a}. The decoders utilize a fully convolutional architecture with a total of 10 convolutional layers. After every two layers the decoder feature maps are nearest-neighbor upsampled and concatenated with encoder features from the same resolution (i.e., skip connections). The depth output is computed at four resolutions using a convolutional layer generating a sigmoid output $\bm{\sigma}_t = (\sigma_{t,i})$.} The depth map is then obtained by $\frac{1}{a \sigma_{t,i} + b}$, where $a$ and $b$ constrain the depth to the range $\left[0.1, 100\right]$. The segmentation output is computed only at the highest resolution by applying a softmax function to the $|\mathcal{S}|$ output feature maps. \revision{single forward pass}{Accordingly, during evaluation, depth and segmentation outputs can be generated in a single forward pass using a single RGB image as input.}
\par
\textbf{Multi-Task Training Details}:
Our multi-task network is trained by employing the Cityscapes training set for the segmentation task and the KITTI training set for the depth task. \revision{mixed batches}{As proposed in \cite{Klingner2020a},} in each training step, we use mixed batches with 6 images from each dataset, i.e., in total 12 images in each batch. \revision{image resolution}{The KITTI and Cityscapes images are resized to resolutions of $192\times 640$ and $512\times 1024$, respectively. The Cityscapes images are then randomly cropped to a resolution of $192\times 640$.} All images are processed by the multi-task network, while the losses are only applied on the task-specific images. We train all networks for 40 epochs using the Adam optimizer~\cite{Kingma2015} with a learning rate of $10^{-4}$, which is reduced to $10^{-5}$ after $30$ epochs. 

\section{Experimental Evaluation}
\label{sec:experiments}

Here, we first analyze our edge-consistent multi-task training. Afterwards, we justify our detection method's design. Finally, we evaluate our method with various perturbations.

\begin{figure*}[t]
	\centering
	\vspace{-0.3cm}
	\subfloat[Example image under perturbation\label{fig:qualitative_images} ]{\includegraphics[width=0.45\textwidth]{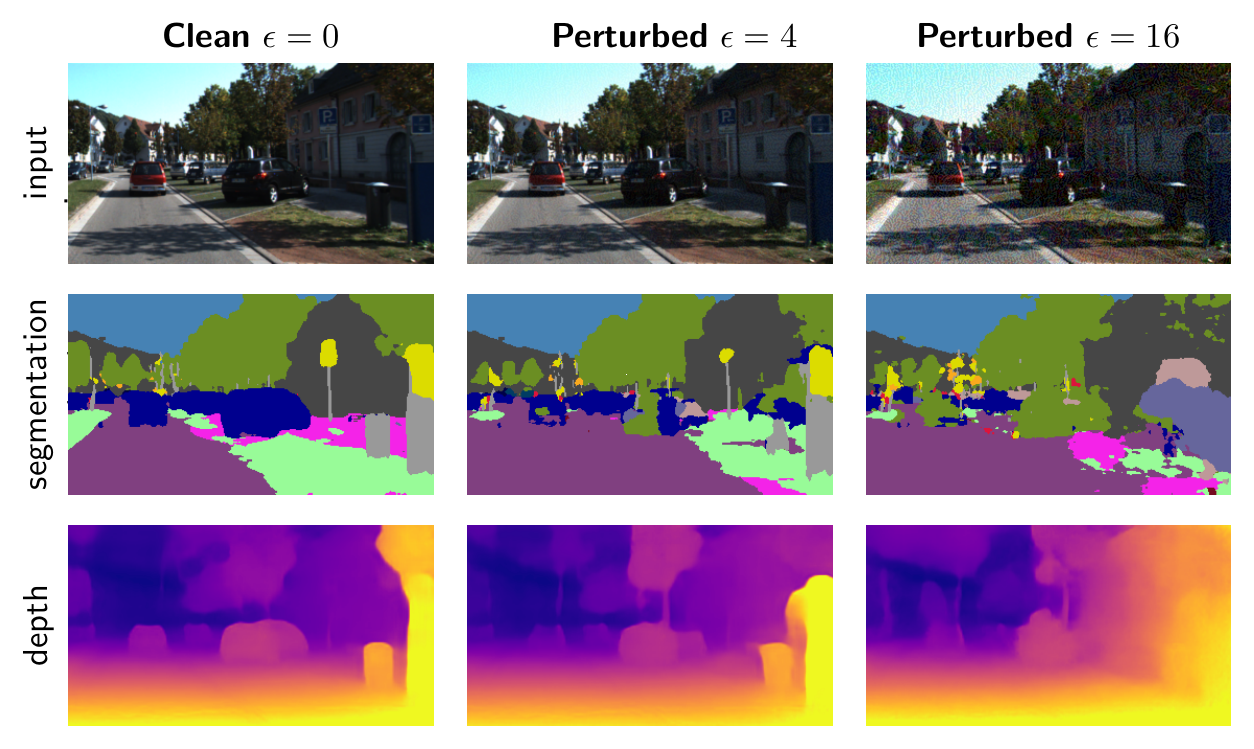}}
	\hspace{0.6cm}
	\subfloat[Top: histogram over consistencies, bottom: ROC curves\label{fig:qualitative_histogramm} ]{\includegraphics[width=0.45\textwidth]{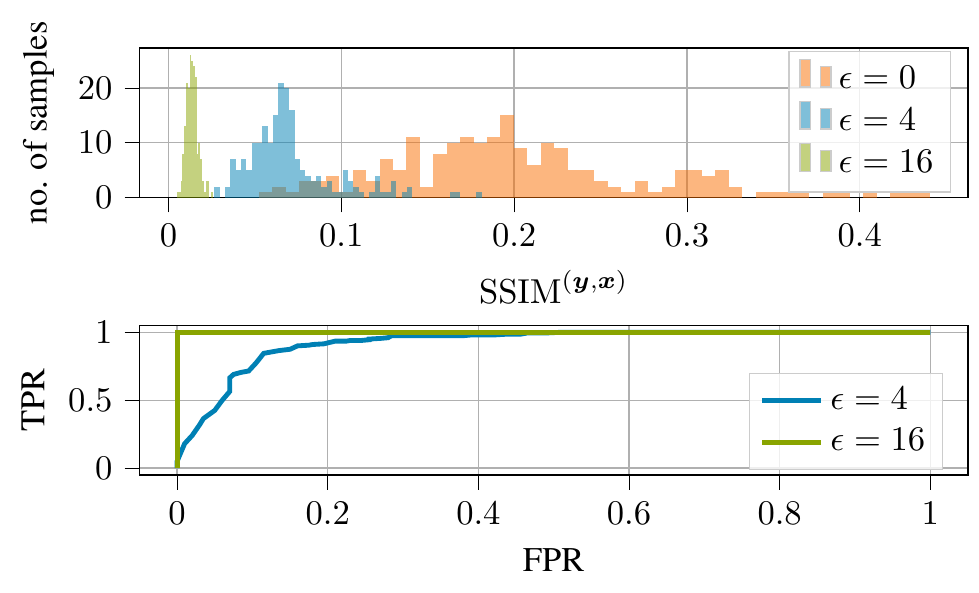}}
	\caption{\textbf{Qualitative results}. We show an example image under an FGSM adversarial attack of different perturbation strengths and corresponding outputs (left). Further, we show a histogram of all samples in the KITTI-2015 validation set over the consistencies between RGB image and segmentation output under different perturbation strengths (right, top) and ROC curves for the detection performance for $\epsilon=4$ and $\epsilon=16$ (right, bottom).}
	\vspace{-0.7cm}
	\label{fig:qualitative}
\end{figure*} 

\subsection{Prediction Performance}

In Table~\ref{tab:prediction_performance}, we evaluate the prediction performance of our method on the KITTI-2015 dataset as it is well-established and provides readily available labels for both depth and segmentation tasks. To show the effect of multi-task training, we first train single-task baselines (only segmentation, only depth). As can be seen, the multi-task model (MT) outperforms both single-task baselines on their respective tasks and provides the capability to compute all three edge consistencies required for our detection method. Adding our proposed edge-consistency loss (\ref{eq:edge_consistency_loss}) (Ours), even increases the segmentation performance slightly to 51.6\%, while keeping the depth performance at a consistently high level. \revision{}{Compared to other monocular approaches~\cite{Klingner2020a, Liu2019a, Yang2018c}, we report a similar or better performance. While the stereo approach from Zhu~et~al.~\cite{Zhu2020} should be better by nature, we can even outperform this approach in some metrics (e.g., in $\delta < 1.25$).}
\par
To facilitate a better adversarial perturbation detection \revision{motivation for edge consistency}{(but not necessarily better single-task performances)}, we are interested in increasing the edge consistency of clean images and respective outputs ($\overline{\mathrm{SSIM}}^{(\bm{m},\bm{x})}$, $\overline{\mathrm{SSIM}}^{(\bm{x},\bm{d})}$, $\overline{\mathrm{SSIM}}^{(\bm{m},\bm{d})}$). \revision{motivation for edge consistency loss}{The main idea is that a better initial edge consistency after training is more sensitive when being exposed to adversarial perturbations during evaluation, therefore being able to indicate such perturbations.} We compute the consistency $\mathrm{SSIM}^{(k,\kappa)}_t$ for each data sample and average over all validation set images (right hand side of Table~\ref{tab:prediction_performance}). We choose the SSIM metric as it does not only compare single pixel values but rather computes the similarity of image patches. 
\par 
First, we employ the commonly used smoothness loss between the depth image and the RGB image from Godard~et~al.~\cite{Godard2017}, however, the loss does not seem to improve our edge consistency between RGB image and depth output as we observe no improvement in the $\overline{\mathrm{SSIM}}^{(\bm{x},\bm{d})}$ metric. We therefore propose a loss function directly enforcing the consistency w.r.t. the SSIM metric between input and both output modalities, i.e., our novel edge consistency loss (ECL) defined in (\ref{eq:edge_consistency_loss}). In the three MT + ECL experiments, we apply the loss only to \textit{one} of the three consistencies, e.g., only to the RGB image and segmentation output in the MT + ECL$(\bm{y},\bm{x})$ experiment. We use a rather strong weighting factor of $\mu=0.01$ for these experiments to strongly enforce consistency. However, we observe that an improvement in one of the consistency measures usually comes at the price of a decreased consistency in another measure. The strength of this effect scales with $\mu$ but was observable in all experiments in this direction. In our full approach, we therefore apply the ECL loss on all three consistencies as described by (\ref{eq:edge_consistency_loss}) and obtain improved consistency in all three measures ($\mu=0.003$ was optimized on the KITTI validation set).
  
\subsection{Detection Design Ablation}

Insights into our method can be gained from Fig.~\ref{fig:qualitative}. We choose an attack design, where only the segmentation task is attacked. We later confirm that additionally attacking the depth task even improves the detection. On the left hand side, we see that the output performance decreases with increasing perturbation strength $\epsilon$. Accordingly, the consistencies of the edges extracted from input and both outputs also decrease, as shown for the consistency between the RGB image and the segmentation output in Fig.~\ref{fig:qualitative}, top right. We observe a distribution shift for adversarial perturbations of strength $\epsilon=4$ and an even stronger shift for $\epsilon=16$, indicating that with a well-chosen threshold the adversarially perturbed images can be detected. In the bottom right part of Fig.~\ref{fig:qualitative}, we confirm this by the ROC curves of our detector under two perturbation strengths. For $\epsilon=4$, we see that for a false positive rate (FPR) of 20\% an about 90\% true positive rate (i.e., detection rate) is achieved. For $\epsilon=16$ it is even possible to choose a threshold such that the FPR is at 0\% while the TPR is at $100\%$, i.e., a perfect detector. For all subsequent evaluations, we fix the FPR to 5\% as a high FPR would generate too many false alarms for practical use cases.
\par
\begin{table}[t]
  \centering
  \vspace{0.2cm}
  \caption{\textbf{Detection performance ablation} for the FGSM adversarial attack on the KITTI-2015 and Cityscapes validation sets for our proposed model (\textbf{Ours}), our model without the edge-consistency loss (w/o ECL), our model where edge consistency (\ref{eq:detection_single}) is measured by the MAE instead of SSIM, and our model with binary edges (BE) where RGB image edges (\ref{eq:rgb_edges}) and depth edges (\ref{eq:depth_edges}) are binarized by setting the top 5\% gradient values to $1$ and the rest to $0$ (segmentation edges (\ref{eq:seg_edges_binary}) are binary anyway). We report the detection rate $\mathrm{TPR}_{5\%}$ $[\%]$ for various perturbation strengths as well as the average detection rate across all considered perturbation strengths. \revision{VGG-16 network}{We further show test results for the reimplemented approach of Hendrycks~et~al.~\cite{Hendrycks2017}, for a variant of our model using the \texttt{VGG-16} network architecture instead of \texttt{ResNet-18}, and for models using either only the segmentation consistency (SC) or only the depth consistency (DC) for detection.} Best results on each dataset in boldface, second best underlined.}
  \resizebox{\columnwidth}{!}{
  \setlength{\tabcolsep}{4.1pt}
  \begin{tabular}{c|l|cccccc|c}
  \multirow{2}{*}{\!\!$\mathrm{TPR}_{5\%}$\!\!} & & \multicolumn{6}{c|}{Perturbation strength $\epsilon$} & \multirow{2}{*}{Average}\\
   & Method & 1 & 2 & 4 & 8 & 16 & 32\\
  \hline
  \multirow{4}{*}{\rotatebox{90}{KITTI}} & Ours with BE \stz & \textbf{8} & 	\underline{9} & 	12 & 	38 & 	\underline{99} & 	\textbf{100} & 	44 \\
   & Ours with MAE & \underline{6} & 	\textbf{8} & 	15 & 	58 & 	\textbf{100} & 	\textbf{100} & 	48 \\
   & Ours w/o ECL & \textbf{8} & 	\underline{9} & 	\underline{35} & 	\underline{70} & 	\underline{99} & 	\underline{97} & 	\underline{53} \\
   & \textbf{Ours}  & \underline{6} & 	\textbf{11} & 	\textbf{42} & 	\textbf{97} & 	\textbf{100} & 	\textbf{100} & 	\textbf{59} \\
  \hline
  \multirow{8}{*}{\rotatebox{90}{Cityscapes}} &  \revision{red}{Hendrycks~et~al.~\cite{Hendrycks2017}}\stz & \revision{red}{\underline{10}} & 	\revision{red}{22} & 	\revision{red}{50} & 	\revision{red}{\underline{92}} & 	\revision{red}{\underline{99}} & 	\revision{red}{\textbf{100}} & 	\revision{red}{62} \\ 
  \cline{2-9}
  & Ours with BE & 6 & 	8 & 	27 & 	86 & 	\textbf{100} & 	\textbf{100} & 	54 \\
  & Ours with MAE & 5 & 	11 & 	45 & 	\textbf{100} & 	\textbf{100} & 	\textbf{100} & 	60 \\
  & Ours w/o ECL & \underline{10} & 	27 & 	82 & 	\textbf{100} & 	\textbf{100} & 	\textbf{100} & 	70 \\
  & \textbf{Ours}  & 9 & 	\textbf{43} & 	\textbf{100} & 	\textbf{100} & 	\textbf{100} & 	\textbf{100} & 	\textbf{75} \\
  \cline{2-9}
  & \revision{red}{\textbf{Ours} with VGG-16}\stz & \revision{red}{\textbf{23}} & 	\revision{red}{\underline{36}} & 	\revision{red}{87} & 	\revision{red}{\textbf{100}} & 	\revision{red}{\textbf{100}} & 	\revision{red}{\textbf{100}} & 	\revision{red}{\underline{74}} \\
  & \revision{red}{\textbf{Ours} only DC} & \revision{red}{8} & 	\revision{red}{35} & 	\revision{red}{97} & 	\revision{red}{\textbf{100}} & 	\revision{red}{\textbf{100}} & 	\revision{red}{\textbf{100}} & 	\revision{red}{73} \\
  & \revision{red}{\textbf{Ours} only SC} & \revision{red}{0} & 	\revision{red}{4} & 	\revision{red}{\underline{98}} & 	\revision{red}{\textbf{100}} & 	\revision{red}{\textbf{100}} & 	\revision{red}{\textbf{100}} & 	\revision{red}{67} \\
  \end{tabular}}
  \vspace{-0.65cm}
  \label{tab:ablation_detection}
\end{table}

To give more insights into our detector design, we provide several ablations in Table~\ref{tab:ablation_detection}. \revision{efficacy of ECL, benefit of initial consistency}{We show that our approach outperforms a model trained without our edge consistency loss (ECL), confirming the efficacy of our ECL loss and showing that a slightly higher initial consistency on clean images (cf. Tab.~\ref{tab:prediction_performance}) clearly supports our detection method (cf. Tab.~\ref{tab:ablation_detection}).} We further show two variants of our detection method: First, we replace the SSIM consistency calculation with a simple mean absolute error (MAE) difference. The decreased performance of this model shows that it is important to compare the structure of the extracted edges. Second, we replaced the gradient calculation by a binary edge extraction (BE), where the top 5\% gradient values are set to $1$ and all other gradient values are set to $0$. The worse performance of this model shows that gradient changes on a continuous scale are an important factor for our adversarial perturbation detection. These properties generalize well to the Cityscapes dataset (also assuming a 5\% FPR).
\par
\revision{VGG-16, single-task and comparison to baseline, real-time}{To verify our method's efficacy and generalizability, we provide further test results on the Cityscapes dataset in Table~\ref{tab:ablation_detection}. First, we reimplement the approach of Hendrycks~et~al.~\cite{Hendrycks2017} and transfer it to the task of semantic segmentation. Our method clearly outperforms this baseline approach. Second, we show that our method can be applied independently of the used network architecture, which is shown by the high detection rate when using the \texttt{VGG-16} network architecture instead of \texttt{ResNet-18}. Third, we show that a variant of our method could also be applied to single-task networks for segmentation or depth. Here, we can only make use of one consistency measure instead of three, either the segmentation consistency (``\textbf{Ours} only SC'') or the depth consistency (``\textbf{Ours} only DC''). Both approaches yield high detection rates, showing the suitability of our approach also for single-task networks. Finally, the multi-task network supplemented by our detection method can be operated at \SI{20}{fps}, where the runtime of our method is negligible compared to the forward pass of the network. This result shows the real-time capability of our method on our used \texttt{NVIDIA Tesla P100} graphics card.}
\par
\begin{figure}[t]
	\centering
	\vspace{0.2cm}
	\includegraphics[width=1.0\linewidth]{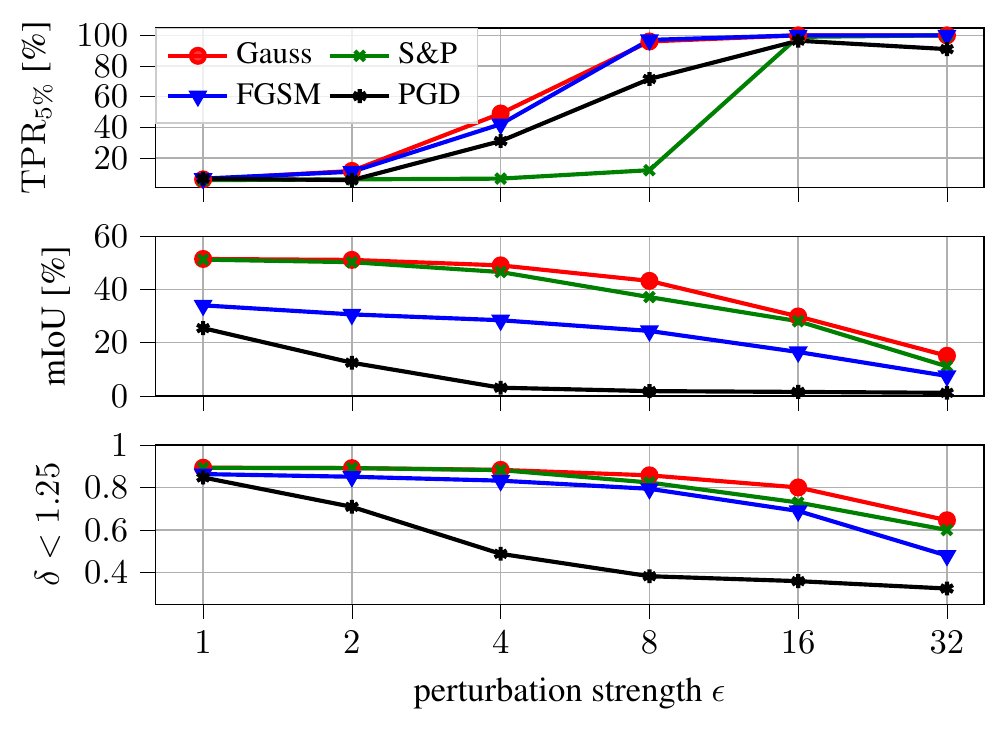}
	\caption{\textbf{Detection performance} measured by detection rate $\mathrm{TPR}_{5\%}$ and \textbf{prediction performance} for semantic segmentation and depth estimation measured by the mIoU and the $\delta<1.25$ metric, respectively, for different perturbation types and strengths $\epsilon$ on the KITTI-2015 validation set.}
	\vspace{-0.5cm}
	\label{fig:performance_detection}
\end{figure}

Next, we are interested to see how the prediction performance of both tasks and the detection performance relate under an increasing perturbation strength $\epsilon$, which we show in Fig.~\ref{fig:performance_detection}. We observe that while the prediction performance of both tasks decreases, the detection rate increases, which is the desired behavior. \revision{limitations}{However, there are still some edge cases (e.g., the PGD attack at strength $\epsilon\leq 4$), which can fool the network to a large degree while not being detected by our method. This seems to be a limitation of detection methods in general as \cite{Hendrycks2017} also exhibits this limitation (cf.~Tab.~\ref{tab:ablation_detection}). Also, since our method detects adversarial perturbations on the entire image, it might show limited performance if perturbations occur only locally, which could, however, potentially be mitigated by applying our detector (\ref{eq:detection_single}) locally on image patches to deliver multiple local results.} Still, all perturbations are reliably detected for $\epsilon\geq 8$, while smaller perturbations could potentially be defended by more robust training methods \cite{Bai2021}. \textit{We conclude that our method provides an upper bound $\epsilon^{\mathrm{max}} = 4...8$ to the perturbation strengths a network needs to be robust against.} 

\subsection{Detection Performance}

\begin{table}[t]
  \centering
  \vspace{0.3cm}
  \caption{\textbf{Detection performance of our method across various perturbation types} on the KITTI-2015 and Cityscapes validation sets. We report the detection rate $\mathrm{TPR}_{5\%}$ $[\%]$ for various perturbation strengths $\epsilon$ as well as the average detection rate across all considered perturbation strengths. $\mathrm{TPR}_{5\%}\geq 90\%$ are printed in boldface.}
  \resizebox{\columnwidth}{!}{
  \setlength{\tabcolsep}{6pt}
  \begin{tabular}{c|l|cccccc}
  \multirow{2}{*}{\!\!$\mathrm{TPR}_{5\%}$\!\!} & \multirow{2}{*}{\shortstack{Perturbation\\ type\quad\quad\quad\quad}} & \multicolumn{6}{c}{Perturbation strength $\epsilon$}\\
   & & 1 & 2 & 4 & 8 & 16 & 32\\
  \hline
  \multirow{6}{*}{\rotatebox{90}{KITTI}} & Gaussian noise\stz & 6 & 	12 & 	49 & 	\textbf{96} & 	\textbf{100} & 	\textbf{100} \\
   & S \& P noise & 5 & 	6 & 	6 & 	12 & 	\textbf{99} & 	\textbf{100} \\
   & FGSM~\cite{Goodfellow2015} & 6 & 	11 & 	42 & 	\textbf{97} & 	\textbf{100} & 	\textbf{100} \\
   & BIM~\cite{Kurakin2017} & 6 & 	7 & 	28 & 	74 & 	\textbf{98} & 	\textbf{99} \\
   & PGD~\cite{Madry2018} & 6 & 	5 & 	31 & 	71 & 	\textbf{96} & 	\textbf{91} \\
   & O-PGD~\cite{Bryniarski2022} & 6 & 	9 & 	26 & 	62 & 	\textbf{99} & 	\textbf{100} \\
  \hline
  \multirow{6}{*}{\rotatebox{90}{Cityscapes}} & Gaussian noise\stz & 11 & 	49 & 	\textbf{99} & 	\textbf{100} & 	\textbf{100} & 	\textbf{100} \\
   & S \& P noise & 4 & 	4 & 	11 & 	65 & 	\textbf{100} & 	\textbf{100} \\
   & FGSM~\cite{Goodfellow2015} & 9 & 	43 & 	\textbf{100} & 	\textbf{100} & 	\textbf{100} & 	\textbf{100}  \\
   & BIM~\cite{Kurakin2017} & 10 & 	38 & 	86 & 	\textbf{99} & 	\textbf{100} & 	\textbf{100} \\
   & PGD~\cite{Madry2018} & 10 & 	37 & 	85 & 	\textbf{100} & 	\textbf{100} & 	\textbf{99} \\
   & O-PGD~\cite{Bryniarski2022} & 11 & 	37 & 	89 & 	\textbf{100} & 	\textbf{100} & 	\textbf{100} \\
  \end{tabular}}
  \vspace{-0.2cm}
  \label{tab:different_attacks}
\end{table}

\begin{table}[t]
  \centering
  \caption{\textbf{Multi-task attack design ablation} for the PGD adversarial attack on our method on the KITTI-2015 and Cityscapes validation sets. We attack only the segmentation task (S), the depth task (D), both tasks at once (SD). We also show an attack on both tasks (SD) \textit{and} the detector by additionally enforcing edge-consistency by the ECL loss (\ref{eq:edge_consistency_loss}), where the number represents the chosen value for $\tilde{\mu}$ in (\ref{eq:multi_task_attack_detection}). We report the detection rate $\mathrm{TPR}_{5\%}$ $[\%]$ for various perturbation strengths as well as the average detection rate across all considered perturbation strengths. $\mathrm{TPR}_{5\%}\geq 90\%$ are printed in boldface.}
  \resizebox{\columnwidth}{!}{
  \setlength{\tabcolsep}{5pt}
  \begin{tabular}{c|l|cccccc}
  \multirow{2}{*}{\!\!$\mathrm{TPR}_{5\%}$\!\!} & & \multicolumn{6}{c}{Perturbation strength $\epsilon$}\\
   & Method & 1 & 2 & 4 & 8 & 16 & 32\\
  \hline
  \multirow{5}{*}{\rotatebox{90}{KITTI}} & S\stz & 6 & 	5 & 	31 & 	71 & 	\textbf{96} & 	\textbf{91} \\
   & D & 5 & 	5 & 	16 & 	64 & 	\textbf{100} & 	\textbf{100} \\
   & SD & 6 & 	6 & 	26 & 	74 & 	\textbf{98} & 	\textbf{93} \\
   & SD + ECL($0.01$) & 5 & 	7 & 	28 & 	76 & 	\textbf{99} & 	\textbf{93} \\
   & SD + ECL($1$) & 5 & 	8 & 	28 & 	77 & 	\textbf{96} & 	\textbf{94} \\
   \hline
   \multirow{5}{*}{\rotatebox{90}{Cityscapes}} & S\stz & 10 & 	37 & 	85 & 	\textbf{100} & 	\textbf{100} & 	\textbf{99} \\
   & D & 6 & 	23 & 	75 & 	\textbf{100} & 	\textbf{100} & 	\textbf{100} \\
   & SD & 9 & 	39 & 	84 & 	\textbf{100} & 	\textbf{100} & 	\textbf{99} \\
   & SD + ECL($0.01$) & 9 & 	36 & 	85 & 	\textbf{99} & 	\textbf{100} & 	\textbf{99} \\
   & SD + ECL($1$) & 11 & 	38 & 	83 & 	\textbf{99} & 	\textbf{100} & 	\textbf{100} \\
  \end{tabular}}
  \vspace{-0.5cm}
  \label{tab:ablation_attack}
\end{table}

We find that the detection performance generalizes well across a range of adversarial perturbations, see results for FGSM, BIM, and PGD in Table~\ref{tab:different_attacks}, and generalizes well to other datasets as can be seen by the strong results on the Cityscapes dataset in Table~\ref{tab:different_attacks}. In particular, our method is able to handle the strong O-PGD attack \cite{Bryniarski2022}, which implicitly optimizes at fooling our detection method. \revision{o-pgd shows superiority}{Our method's efficacy is underlined by the good detection results on the O-PGD attack, as this attack has been shown to be capable of fooling several state-of-the-art detection methods on the image classification task.}
\par 
Finally, we show how variants of our adversarial attack affect the detection performance in Table~\ref{tab:ablation_attack}. Only attacking the depth task (D) yields similar detection performance as only attacking the segmentation task (S). Attacking both tasks (SD) as outlined in (\ref{eq:multi_task_attack}) even improves detection (cf.~Table~\ref{tab:different_attacks}), as there is stronger edge inconsistency due to both outputs being strongly impaired. Enforcing edge consistency during optimization of the adversarial perturbation via our ECL loss (SD+ECL experiments, cf.~(\ref{eq:multi_task_attack_detection})) does not really affect detection performance. \revision{strength of not being fooled}{The reason could be that increasing the consistency between input image and network outputs, and decreasing the network performance are strongly conflicting goals, which are hard to optimize. This is even supported by the theoretical analysis in \cite{Mao2020}, which shows that attack optimization becomes more difficult with an increasing number of tasks. We therefore conclude that our detection method exhibits a clear strength because it cannot be fooled even if the attacker has white box knowledge. }

\section{Conclusions}
\label{sec:conclusion}

In this paper, we presented an adversarial perturbation detection method based on multi-task perception of the two complex vision tasks semantic segmentation and depth estimation. By detecting low edge consistency between input and outputs, we achieve up to 100\% detection rate on strong adversarial perturbations, while 5\% clean images are wrongfully classified as perturbed. We further show that our detection method benefits from a consistency-enforcing training and that even an attacker with white-box knowledge about the model properties cannot easily fool our detection method. Since our detection method reliably handles strong perturbations, it puts a natural upper limit to a perturbation strength the network must be robust against. Future work might combine our detection method with a more robust training or adversarial defenses to reduce the negative impact of smaller perturbations on the network's performance.

\bibliographystyle{IEEEtran}
\bibliography{IEEEabrv,bib/ifn_spaml_bibliography}

\end{document}